\documentclass[10pt,twocolumn,letterpaper]{article}

\usepackage{cvpr}              

\usepackage[pagebackref,breaklinks,colorlinks]{hyperref}

\def\paperID{*****} 
\def\confName{CVPR}
\def\confYear{2024}

\title{The Solution for the GAIIC2024 RGB-TIR object detection Challenge}

\author{Xiangyu Wu\textsuperscript{1},
Jinling Xu\textsuperscript{2},
Longfei Huang\textsuperscript{1},
Yang Yang\textsuperscript{1}\thanks{Corresponding author}\\
\textsuperscript{1}{Nanjing University of Science and Technology}
\textsuperscript{2}{APe counseling}\\
yyang.@njust.edu.cn
}

\begin{document}
\maketitle

\begin{abstract}
This report introduces a solution to The task of RGB-TIR object detection from the perspective of unmanned aerial vehicles. Unlike traditional object detection methods, RGB-TIR object detection aims to utilize both RGB and TIR images for complementary information during detection. The challenges of RGB-TIR object detection from the perspective of unmanned aerial vehicles include highly complex image backgrounds, frequent changes in lighting, and uncalibrated RGB-TIR image pairs. To address these challenges at the model level, we utilized a lightweight YOLOv9 model with extended multi-level auxiliary branches that enhance the model's robustness, making it more suitable for practical applications in unmanned aerial vehicle scenarios. For image fusion in RGB-TIR detection, we incorporated a fusion module into the backbone network to fuse images at the feature level, implicitly addressing calibration issues. Our proposed method achieved an mAP score of 0.516 and 0.543 on A and B benchmarks respectively while maintaining the highest inference speed among all models.
\end{abstract}

\section{Introduction}
\label{sec:intro}
Based on drones, vehicle detection plays a crucial role in intelligent urban traffic management and disaster relief efforts. Drones equipped with cameras can capture images with a wide field of view, which is advantageous for capturing ground targets. However, due to highly complex backgrounds and frequent changes in lighting, airborne-based object detection remains an active and challenging task in the field of computer vision \cite{YuanW24, YuanWW22}.

\begin{figure}[t]
	\centering
	\includegraphics[width=\linewidth]{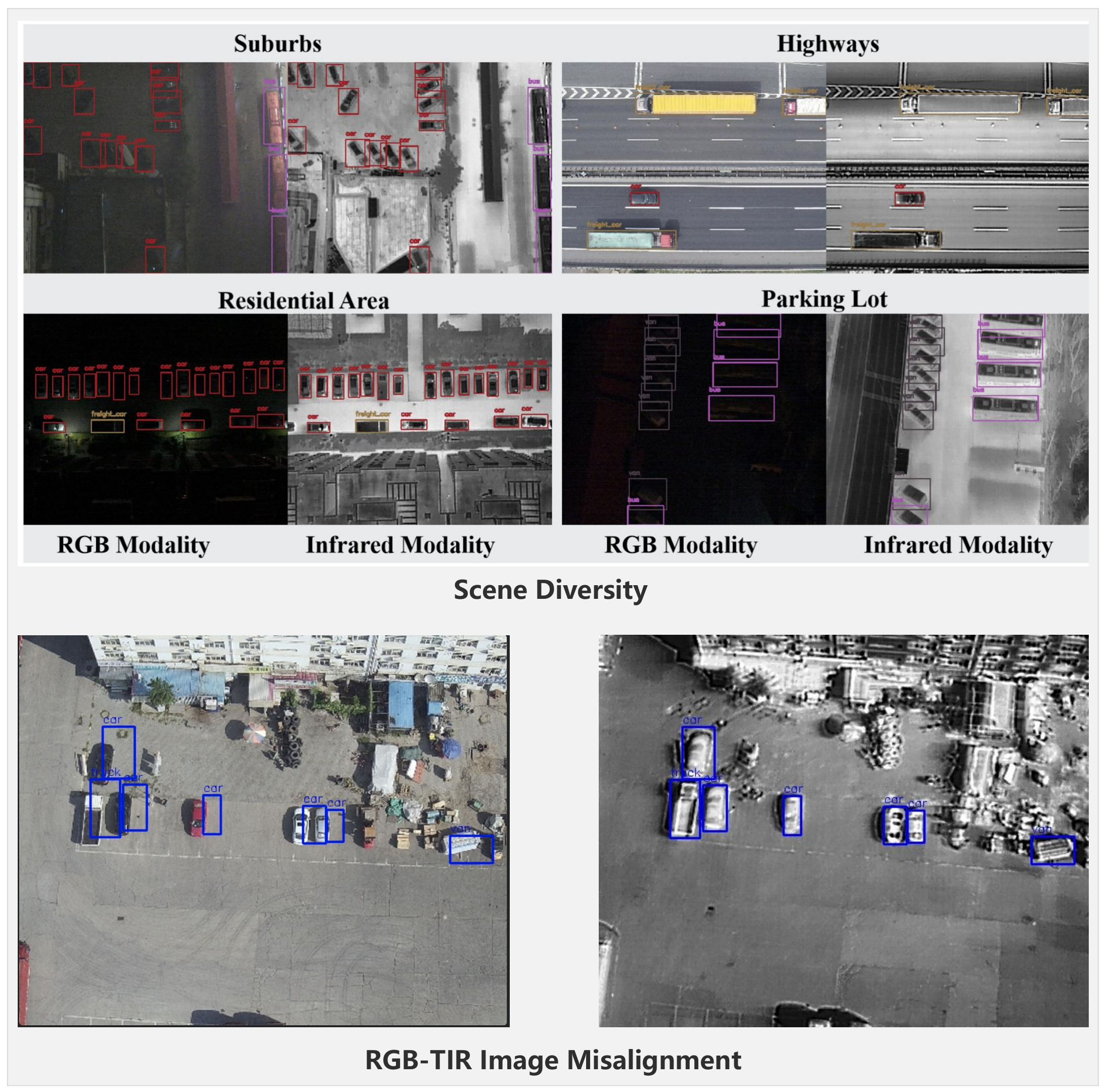}
	\caption{The data comprises various scenes and environments. There are significant differences in the quality and clarity between RGB and TIR images, with notable noise issues. Variations in imaging time and the operational states of different vehicles lead to slight discrepancies in the positions of corresponding vehicles in paired images.}\label{fig: dataset}
\end{figure}

With the development and application of multimodal technology\cite{YangGLLLY24, YangBGZYY23, abs-2110-11767, YangYBZZGXY23, YangWZL019}, considering the robustness of infrared cameras in all-day imaging, a challenge arises in integrating infrared images to complement visible light modalities, forming RGB-TIR image pairs. However, RGB-IR object detection faces challenges in modal calibration errors\cite{YuanW24}, stemming from RGB-IR image pairs captured by sensors with different fields of view and imaging times, resulting in potential asynchrony in the positions of moving objects. Additionally, the RGB-TIR target detection dataset from the drone's perspective covers various scenes including urban roads, rural areas, residential areas, parking lots, and spans from day to night, as shown in Figure \ref{fig: dataset}.

To tackle these challenges, we opted for the lightweight YOLOv9 model\cite{abs-2402-13616}, which better suits practical application scenarios. Additionally, we introduced external datasets and expanded the model with multi-level auxiliary branches to enhance its robustness. Furthermore, by incorporating a model fusion module, we conducted image fusion at the feature level to implicitly address image misregistration issues and improve model performance. We also integrated powerful data augmentation techniques to enhance model generalization and cross-domain robustness further.

\begin{figure*}[!h]
	\centering
	\includegraphics[width=350pt]{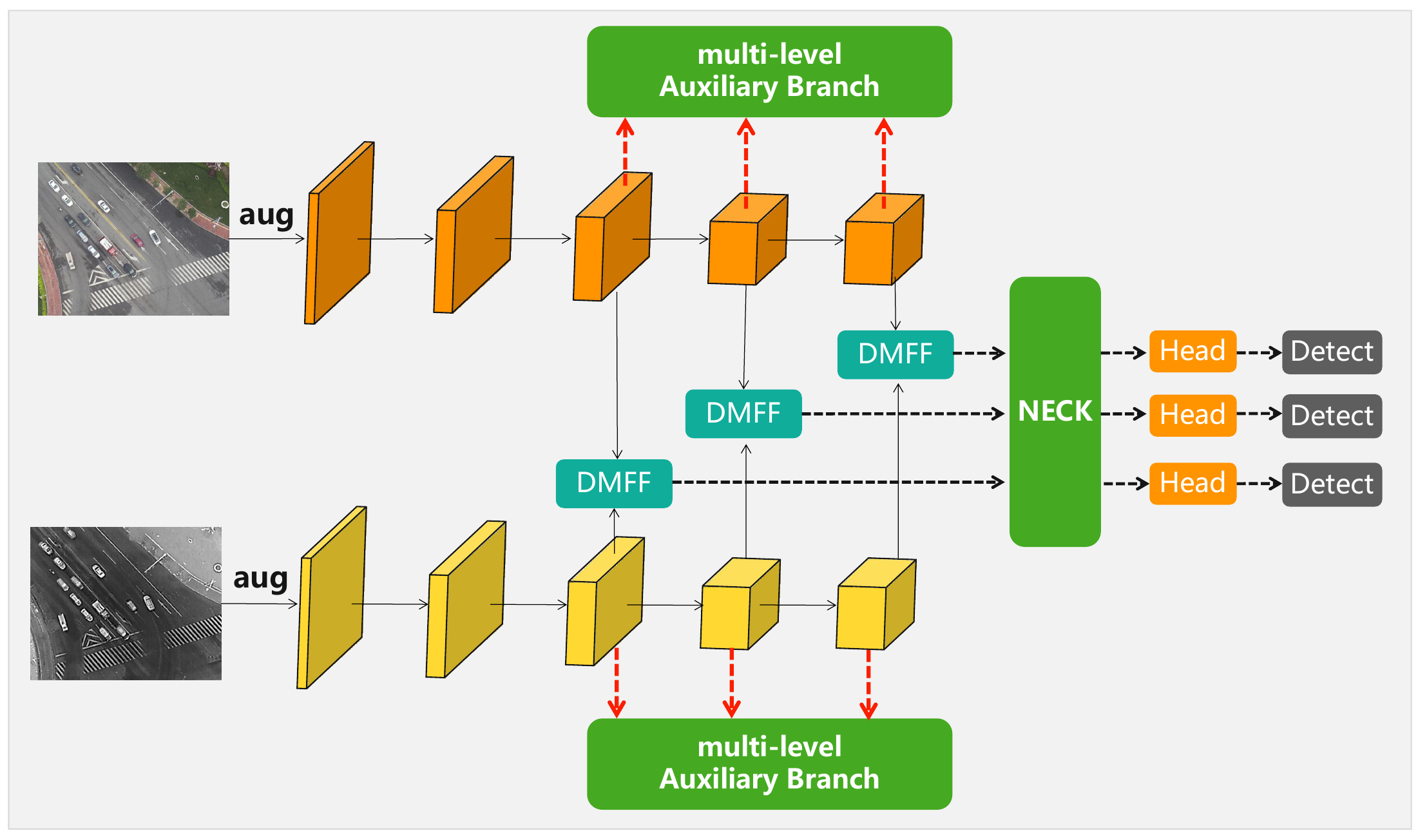}
	\caption{Overall Architecture. Our method consists of three main components: extended multi-level auxiliary branches, image feature-level fusion, and data augmentation and model ensemble. We apply different data augmentation techniques to different modal images and perform image fusion at three scales.}\label{fig: overview}
\end{figure*}

\section{Related Work}
\label{sec:Related}
\subsection{Object Detection}

With the advancement of deep learning technologies, object detection techniques have become increasingly mature. Currently, the mainstream object detectors include the YOLO series\cite{RedmonDGF16, abs-2402-13616} and the DETR series\cite{CarionMSUKZ20, ZongS023}. \cite{ZongS023} introduced the Co-DETR model, which achieved remarkable results on the COCO dataset\cite{LinMBHPRDZ14}. However, due to the extreme difficulty of applying DETR series detectors to new domains without corresponding domain-pretrained models, the most widely used object detector in practical applications remains the YOLO series. The YOLOv9 model, by introducing PGI and GELAN structures, enhances detection performance while ensuring real-time capabilities.

\subsection{Multi-modal Image Fusion}

Currently, multimodal research and applications encompass various techniques\cite{YangZWLXJ21, YangFZLJ21, 0074ZGGZ22, YangZXYZY21, YangWZX019}, among which image fusion methods play a crucial role. Traditional image fusion approaches rely on manually designed fusion rules and features to integrate image information\cite{VishwakarmaB19, ZhouFMY23}. Deep learning-based methods have emerged as the predominant approach to addressing the challenges of image fusion tasks\cite{abs-2402-16267, ShenCLZFY24}. Currently, image fusion can be categorized into image-level fusion and feature-level fusion. Image-level fusion aims to merge images into a single image before inputting them into the model for training and inference. On the other hand, feature-level fusion involves merging the features of images after they have been inputted into the model through fusion modules. Image-level fusion can potentially distort images and lead to information loss, whereas feature-level fusion enables models to adaptively learn information from different modalities of images.

\section{Methodology}

Our approach is primarily divided into three parts: multi-level auxiliary supervision branch strategy, feature-level image fusion strategy, and data augmentation and model ensemble strategy. The overall framework is illustrated in Figure \ref{fig: overview}.

\begin{figure}[t]
	\centering
	\includegraphics[width=150pt]{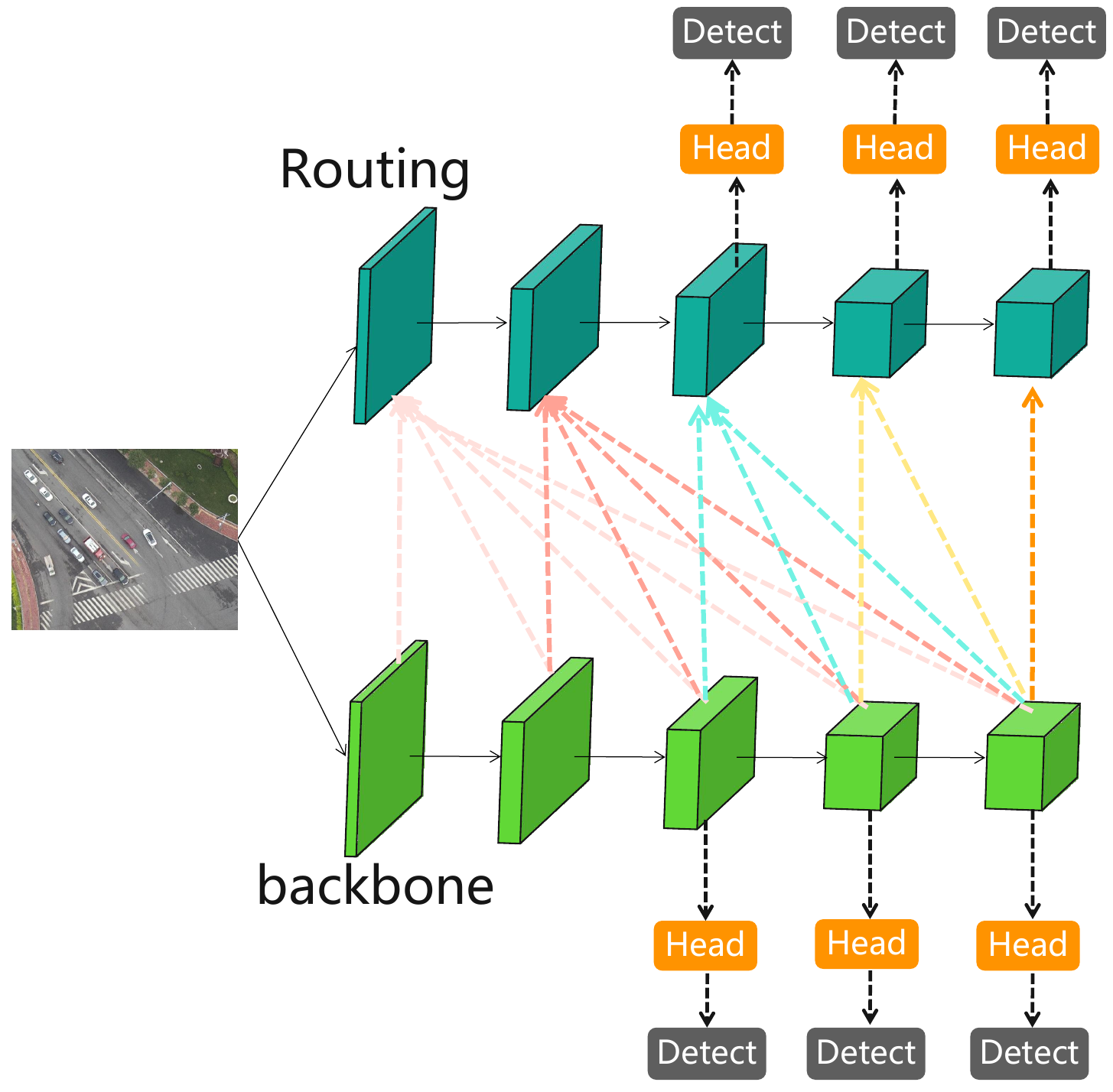}
	\caption{Multi-level Auxiliary Supervision Branch strategy (taking RGB images as an example). }\label{fig: MLASB}
\end{figure}

\subsection{Multi-level Auxiliary Supervision Branch} 

We extended the official YOLOv9 model into a Dual-Backbone structure and incorporated YOLOv9's auxiliary supervision branch strategy. For each modality, we constructed two auxiliary branches: one supervising features before routing and another after routing, aiming to model shallow, high-level, and cross-modal interactions between high and low-level features. This approach allows simultaneous capture of low-level and high-level image characteristics and further learning of interaction information between RGB and TIR modalities across different layers. The multi-level auxiliary supervision branch structure is illustrated in Figure \ref{fig: MLASB}.

\subsection{Feature-level Image Fusion}

Drawing inspiration from the ICAFusion model \cite{ShenCLZFY24}, we adopted a transformer-based cross-modal fusion approach using cross-modal attention for image fusion. This method was integrated into the backbone network of our model, enabling it to autonomously adapt to the characteristics of dual-light images during training. It effectively utilizes information from different modalities and cross-modal interactions, thereby mitigating image misalignment issues. 

\subsection{Data Augmentation and Model Ensemble}

We employ multiple data augmentation techniques tailored to enhance the noise resistance and cross-domain robustness of our model using RGB and TIR images. The data augmentation consists of three parts: (1) developing RGB data augmentation strategies that align with the data characteristics to simulate the distribution of real data; (2) creating cropping and rotation augmentation strategies specific to RGB and TIR data, mimicking challenges in RGB and TIR image registration; (3) devising TIR image data augmentation strategies that simulate imaging blur issues encountered in practical applications.

\begin{figure}[t]
	\centering
	\includegraphics[width=\linewidth]{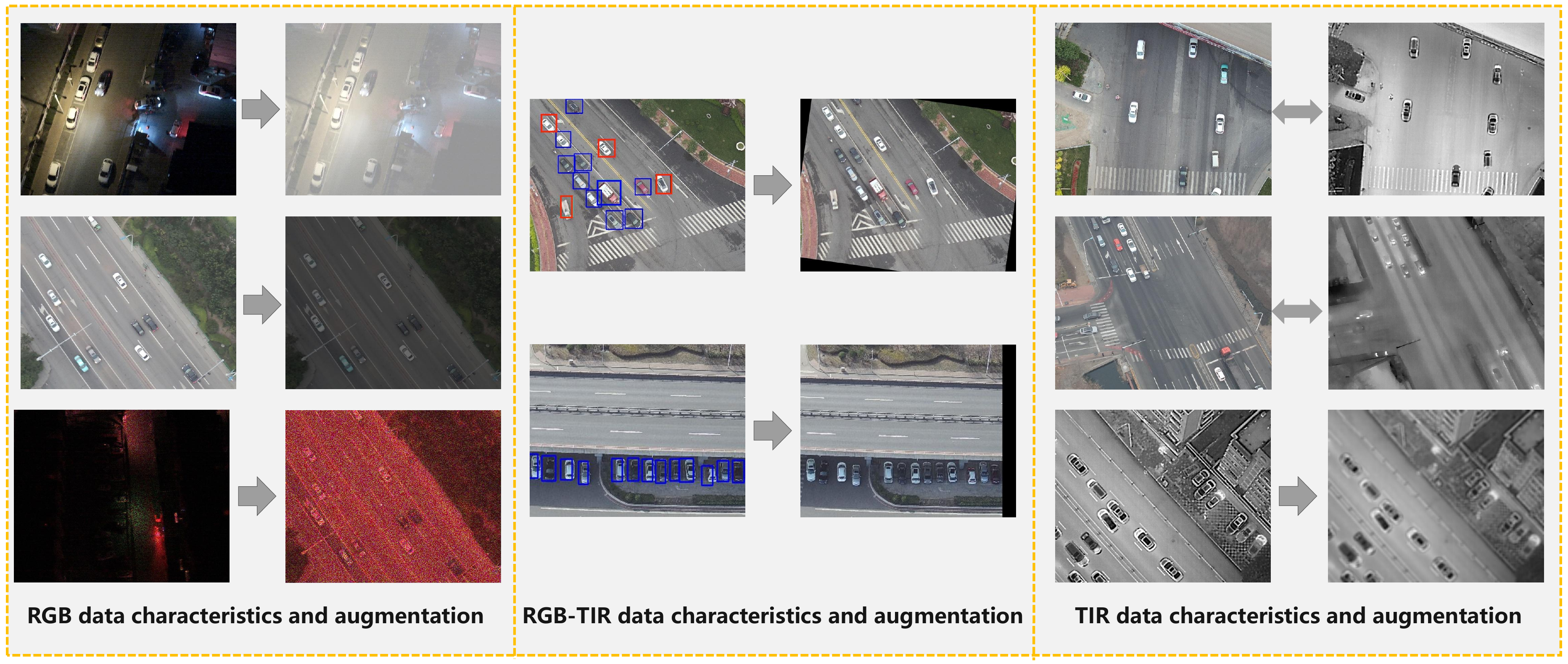}
	\caption{Multi-level Auxiliary Supervision Branch strategy (taking RGB images as an example). }\label{fig: data-aug}
\end{figure}

Specifically, both RGB and infrared images undergo Mosaic augmentation and random noise augmentation. For RGB images, random brightness augmentation is applied, while for TIR images, random edge enhancement, and random blur augmentation are used. Additionally, either cropping or rotation augmentation is randomly applied to both types of images. The effects of data augmentation are illustrated in Figure \ref{fig: data-aug}.

In terms of model ensemble, we employed a differentiated ensemble approach aimed at maximizing result variance. We implemented various experimental strategies to maximize performance enhancement based on maximizing result disparities.

\section{Experiments}
\subsection{Experiments Setup}

\textbf{Dataset}. We introduced external datasets, namely the DroneVehicle dataset\cite{SunCZH22} and the Visdrone dataset. For the DroneVehicle dataset\cite{zhu2021detection}, we directly utilized the original test set data, comprising 8980 image pairs. Regarding the Visdrone dataset, we employed a cropping strategy that calculated the maximum allowable number of crops for each image based on its longest and shortest sides. We excluded crop results that did not contain any objects, resulting in a total of 15,929 RGB images. Since the Visdrone dataset lacks infrared data, we compensated for the absence of infrared images by converting RGB images to grayscale.

\textbf{Implementation Detail}. We conducted training using 4 A100 GPUs, adhering to the default settings of the YOLOv9 model for all parameters. Regarding data augmentation, images had a 0.3 chance of being rotated with random angles between -5 and 5 degrees. There was also a 0.3 chance of applying horizontal or vertical cropping with random pixel values between -10 and 10. Brightness augmentation was exclusively applied to RGB images, while edge enhancement was applied solely to infrared images. Additionally, either cropping or rotation augmentation was randomly applied to either RGB or infrared images.

\begin{table}[!ht]
\centering
\begin{tabular}{ccc}
\toprule[1.5pt]
\# & \textbf{Method} &  \textbf{mAP} \\ 
\midrule 
1 & YOLOv9 & 0.451 \\
2 & +external data & 0.461 \\
3 & +data augmentation & 0.508 \\
4 & +auxiliary branch & 0.525 \\
5 & +model ensemble & 0.543 \\
\bottomrule[1.5pt]
\end{tabular}
\caption{Ablation experiments.} \label{tab:table-1}
\end{table}

\subsection{Main Result}

Our method achieved an mAP score of 0.543 on the A leaderboard and 0.516 on the B leaderboard. Our FPS is 26/s. The results of the ablation experiments on the A leaderboard are shown in Table \ref{tab:table-1}. The experimental results indicate that data augmentation strategies enhance the model's noise resistance and effectively improve its cross-domain robustness. Incorporating multi-level auxiliary supervision branches solely during the training phase further facilitates the learning of interaction information between modalities. This approach ensures real-time capability while enhancing the algorithm's robustness.

\section{Conclusion} 

We developed multi-level auxiliary supervision branches to capture both low-level and high-level image features, facilitating the learning of interaction information between RGB and TIR modalities. Implemented within the YOLOv9 model for real-time efficiency, this approach enhances algorithmic robustness. For image fusion, we utilized the fusion units from the ICAFusion model, enabling the model to autonomously adapt to dual-light image characteristics during training. This method effectively integrates dual-light images at the feature level, mitigating image misalignment issues.

Furthermore, employing various data augmentation techniques tailored separately for RGB and TIR images enhances the model's noise resistance and improves its generalization on the test set. This approach effectively addresses challenges such as complex backgrounds and frequent lighting variations in images, thereby enhancing the model's cross-domain robustness. Our method successfully tackles the object detection challenges presented by RGB-TIR images captured from a drone's perspective.

{
    \small
    \bibliographystyle{ieeenat_fullname}
    \bibliography{main}
}


\end{document}